\pdfoutput=1
\PassOptionsToPackage{dvipsnames}{xcolor}

\documentclass[11pt]{article}

\usepackage[final]{acl}
\usepackage{comment}

\usepackage{times}
\usepackage{latexsym}
\usepackage{booktabs}
\usepackage{amssymb}
\usepackage{array}
\usepackage[T1]{fontenc}

\usepackage[utf8]{inputenc}

\usepackage{microtype}

\usepackage{inconsolata}

\usepackage{graphicx}
\usepackage{booktabs,multirow}

\usepackage{latexsym}
\newcommand{\cmark}{\textcolor{teal}{\ding{51}}}%
\newcommand{\xmark}{\textcolor{purple}{\ding{55}}}%
\usepackage{xspace}

\usepackage{times}
\usepackage{latexsym}
\usepackage{booktabs}
\usepackage[T1]{fontenc}

\usepackage[utf8]{inputenc}

\usepackage{microtype}
\usepackage{enumitem}
\setlist{nolistsep}

\usepackage{latexsym}
\usepackage{xspace}

\usepackage{tabularx}
\usepackage{booktabs}
\usepackage{geometry}

\usepackage{inconsolata}

\usepackage{graphicx}

\usepackage{adjustbox}
\usepackage{tcolorbox}
\usepackage{float}

\usepackage{latexsym}
\usepackage{booktabs}
\usepackage{graphicx}
\usepackage{amsmath}
\usepackage{cleveref}
\usepackage{pifont}
\usepackage{url}
\usepackage{hyperref}

\title{Knowledge-Graph Based Augmentation \textit{versus} Retrieval Augmented Generation for Cultural-Related Question Answering}

\author{
 \textbf{Pablo Poulenard\textsuperscript{1,2}},
 \textbf{Yannis Karmim\textsuperscript{1,3,4}},
 \textbf{Valentin Barrière \textsuperscript{1,5}}
\\
\\
 \textsuperscript{1}DCC, Universidad de Chile, Santiago, Chile
 \textsuperscript{2}École Polytechnique, Palaiseau, France,
\\
 \textsuperscript{3}Inria, Almanach, Paris, France,
 \textsuperscript{4}Inria Chile, Santiago, Chile,
 \textsuperscript{5}CENIA, Macul, Chile
\\
 \small{
   \textbf{Correspondence:} \href{pablo.poulenard@polytechnique.edu}{pablo.poulenard@polytechnique.edu}
 }
}

\begin{document}
\maketitle
\begin{abstract}

Large language models (LLMs) suffer from a long-tail deficit: culturally specific facts, particularly those concerning underrepresented regions such as Latin America, appear too rarely in pretraining corpora to be reliably memorized. Retrieval-Augmented Generation (RAG) addresses this by grounding generation in external text, but structured alternatives such as Knowledge Graphs (KGs) offer tighter control over what enters the context, along with potential gains in explainability and updatability. We benchmark Graph-RAG against standard RAG on LatamQA, a culturally grounded multiple-choice dataset spanning eight thematic categories. The graphs are built end-to-end from Wikipedia articles with KGGen, a recent open-domain extractor, without manual curation in our main setting. G-Retriever is competitive with RAG and reduces the error of the base LLM by 72\% with a standard KG and 78\% with a benchmark-aware variant, the gap to RAG narrowing further as the graph is oriented toward task-relevant content. The trained projection transfers zero-shot to Portuguese without target-language fine-tuning, indicating multilingual reach. Our code is available
\href{https://github.com/Payblito/KG-vs-RAG-QA}{here}.

\end{abstract}

\section{Introduction and Related Work}

LLMs acquire factual knowledge as a by-product of next-token prediction, so recall reliability scales with pretraining frequency \cite{Mallen2023}. Culturally specific facts, and especially those pertaining to underrepresented regions, appear too rarely to be memorised reliably, and the gap manifests as confabulation rather than abstention. Latin American culture is a canonical instance: despite Spanish being a high-resource language, state-of-the-art LLMs answer questions about Iberian Spanish culture substantially more accurately than equivalent questions about Latin American culture \cite{karmim2026}. Parametric adaptation does not close this gap: continued pretraining on a new distribution induces catastrophic forgetting \cite{Yang2026a}, and supervised fine-tuning is superseded by retrieval methods at scale \cite{Ovadia2024}.

Retrieval-Augmented Generation (RAG) addresses these limitations by grounding generation in an external text store without modifying model weights, and is highly effective in low-frequency factual regimes \cite{Mallen2023}. Its main drawback is token cost: retrieved passages fill the context window, and the practitioner has little control over what enters the prompt. A complementary line of work targets not what is retrieved but how it is reasoned over: \cite{ranaldi-etal-2025-improving-multilingual} improve multilingual RAG by having the model compare and reconcile heterogeneous retrieved passages through dialectic argumentation. Knowledge Graphs (KGs) offer a structured alternative: explicit relational triples are compact, easy to inspect, update, and trace back to sources. Automatic KG construction at scale has recently become tractable with KGGen \cite{mo2025}, which converts raw text into triples via structured LLM prompting and reduces graph sparsity through entity and relation resolution.

G-Retriever \cite{he2024} integrates a GNN-based soft prompt with PCST subgraph retrieval and is the leading graph-augmented LLM pipeline. Its evaluations use ExplaGraphs \cite{saha2021}, SceneGraphs \cite{hudson2019}, and WebQSP \cite{yih2016}: datasets that pair each question with a small, clean, dedicated graph and require multi-hop reasoning, conditions structurally favorable to graph-based methods. We evaluate on a deliberately harder regime: a single large, noisy KG per thematic category with single-hop questions, so that RAG is naturally advantaged and the true cost of converting text into triples can be measured.

\begin{figure*}
    \centering
    \includegraphics[width=1.\linewidth]{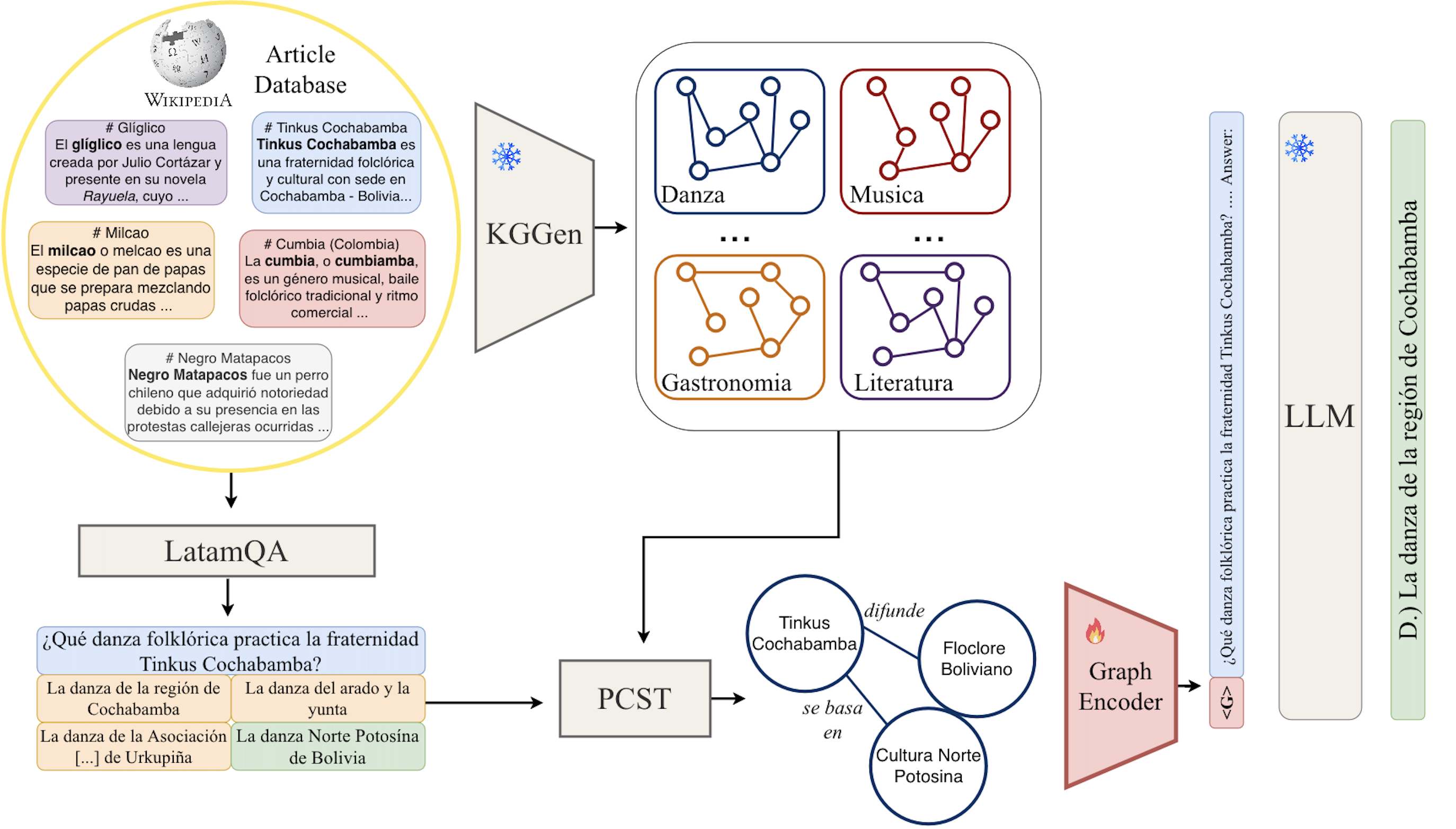}
    \caption{Overview of our method. From 5{,}848 Spanish Wikipedia articles, KGGen \cite{mo2025} extracts one schema-free KG per thematic domain. For a LatamQA \cite{karmim2026} question, G-Retriever \cite{he2024} selects a subgraph via PCST and prepends its encoding \texttt{<G>} to a frozen LLM; the same triples serve as text ($D$) for our RAG and Top-$k$ baselines.}
\vspace{-.3cm}
    \label{fig:overview}
\end{figure*} 

We benchmark Graph-RAG against standard RAG on LatamQA \cite{karmim2026}, a culturally grounded MCQ dataset spanning eight Latin American thematic categories. Our contributions are: \textbf{(i)} a large-scale application of the recent open-domain extractor KGGen \cite{mo2025} to 5{,}848 Wikipedia articles, evaluated on downstream QA rather than intrinsic extraction metrics; \textbf{(ii)} a systematic comparison of RAG, Top-$k$ triples, and G-Retriever \cite{he2024}; \textbf{(iii)} ablation studies isolating the contributions of graph structure and extraction quality; and \textbf{(iv)} zero-shot multilingual transfer of the trained projection to Portuguese. \Cref{fig:overview} present an overview of our proposed pipeline.

\section{Proposed experimental pipeline}
\label{sec:pipeline}
Our setup combines a culturally grounded QA benchmark, KGs extracted from the
articles it is built from, and a small LM to answer the questions.

\paragraph{Dataset}
We build upon LatamQA \cite{karmim2026}, a multiple-choice benchmark of culturally grounded factual knowledge extracted from Wikipedia articles on the cultures of Latin American countries. We define eight thematic categories from the Wikipedia ontology (\textit{Musica}, \textit{Literatura}, \textit{Cinema}, \ldots, see \Cref{tab:kg-stats}), scrape each country-theme mother category (\textit{e.g.} \textit{Gastronom\'ia de Chile}) recursively up to depth two, and intersect the result with LatamQA, yielding 5{,}848 articles. Each article supports exactly one question, with one correct answer and three distractors grounded in its content, so no question requires composition across articles.

\paragraph{Graph construction}
We rely on KGGen \cite{mo2025}, a recent extractor that predicts entities and relations from raw text via structured LLM prompting, then resolves duplicates by iterative clustering (prompts in Appendix~\ref{app:kggen}). It has so far been evaluated only on English corpora of at most 5M tokens; we apply it to 50M characters of Spanish Wikipedia, a regime in which entity resolution dominates runtime. 
Articles are chunked into 5{,}000-character segments and per-article graphs are aggregated by category with entity and edge resolution, yielding \textbf{one unified KG per category, linked with a central node in one global KG}. Each triple is traced through the pipeline, providing provenance used as retrieval ground truth. We additionally construct a \textit{benchmark-aware} variant per category by injecting entity and relation hints derived from LatamQA questions into the extraction prompt, steering the KG toward task-relevant content. Mistral Small 3.2\footnote{\texttt{Mistral-Small-3.2-24B-Instruct-2506}} is the KGGen backbone.

\begin{table}[t]
\centering
\small
\setlength{\tabcolsep}{4pt}
\resizebox{\columnwidth}{!}{%
\begin{tabular}{lrrrr}
\toprule
Category & \#Articles & \#Chars & \#Nodes & \#Edges \\
\midrule
Música       & 1{,}306 & 14M  & 82k & 32k \\
Literatura   & 1{,}515 & 12M & 75k & 29k \\
Cine         & 1{,}269 & 12M & 73k & 30k\\
Folclore     &   575 &  5M &  39k & 16k\\
Gastronomía  &   388 &  3M &  15k &  9k  \\
Danza        &   378 &  2M &  24k & 9k \\
Pintura      &   277 &  2M &  18k & 6k\\
Artesanía    &   140 &  1M &  10k & 4k \\
\midrule
Global       & 5{,}848 & 51M & 336k & 135k \\
\bottomrule
\end{tabular}
}
\caption{Statistics of the eight thematic KGs, after entity and edge resolution} \vspace{-.3cm}
\label{tab:kg-stats}
\end{table}
  
\paragraph{Language Model}
Our main experiments use \texttt{Qwen2.5-3B-Instruct} \cite{qwen2025}. Measuring
the contribution of external knowledge requires a model that has not already
memorized the target facts, since strong zero-shot performance would leave
little headroom to attribute retrieval gains. At 60.17 zero-shot accuracy,
parametric knowledge does not saturate the benchmark.

\paragraph{Query encoding for retrieval}
All four systems share the same encoder and query format. \texttt{jinaai/jina-embeddings-v3} \cite{sturua2024a}, a multilingual encoder with an 8{,}192-token context window, is used throughout, with its
task-specific LoRA adapters for queries and passages (details in Appendix~\ref{app:emb_query}). The query concatenates the question with its four options: this symmetrically enriches the lexical and semantic signal available to the retriever while preserving the integrity of the task, since no option is privileged at retrieval time. \Cref{sec:methods} describes how each system uses this signal to select context.

\section{Compared retrieval methods}
\label{sec:methods}
All systems below use the same language model and embedding model described in \Cref{sec:pipeline}, and differ
only in the context they retrieve. RAG operates on raw text and serves as our
reference point; the three graph-based systems consume the KG with increasing
use of its structure, from an unordered set of triples to a trained subgraph
encoder.
\paragraph{RAG}
Each article is segmented into chunks of at most 512 tokens with an overlap of
64 tokens, encoded with the same model as the queries. The $k=5$ chunks with
the highest similarity to the query are passed to the reader.  Retrieval is restricted
to the articles of the corresponding category, matching the scope of the KG
used by the graph-based methods. These values were
selected by grid search; the full sweep is reported in
Appendix~\ref{appendix:hyperopti}.

\paragraph{Top-$k$ triples}
This training-free baseline scores every triple independently: each $(s,r,o)$
is encoded, ranked by cosine similarity to the query, and the top-$k$ triples
are verbalized into the prompt. This follows the similarity-based filtering of
KAPING \cite{baek2023} but omits its entity-linking stage, which restricts
candidates to the one-hop neighborhood of the question entities: since our graph
comes from open extraction rather than a canonical knowledge base, no exact
correspondence between question and graph entities is guaranteed. The graph is
treated as an unordered set of triples, making this a natural lower bound for
the structure-aware methods below.

\paragraph{G-Retriever}
G-Retriever \cite{he2024} first extracts a subgraph by solving a
Prize-Collecting Steiner Tree over the KG, using query similarity to edges and
vertices as node prizes and edge costs. A graph encoder\footnote{We use a Graph
Transformer \cite{Yun2019} rather than the Graph Attention Network
\cite{Velickovic2018} of the original work.} maps this subgraph to a single
vector, which an MLP projects into the LM embedding space as a soft prompt.
Both modules are trained end-to-end on a training split to produce the correct
answer. In \Cref{subsec:ablation} we also evaluate a variant replacing the
graph encoder by mean pooling over node embeddings, where topology determines
which nodes are retrieved but is never encoded.

\section{Results and Analysis}
We first compare all four systems on LatamQA~\cite{karmim2026} (\Cref{subsec:main}), then isolate the contribution of each component of G-Retriever(~\Cref{subsec:ablation}), and finally test whether the trained projection transfers to an unseen language (~\Cref{sec:multilingual}).
\subsection{Main comparison}

\paragraph{RAG vs Top-$k$ triples vs G-Retriever}
\label{subsec:main}
Table~\ref{tab:results-per-category} compares all methods against zero-shot (60.17). All retrieval methods substantially outperform the unaugmented model. The training-free Top-$k$ triples baseline (+15.4 pp) confirms that even unordered triples carry discriminative signal, yet it remains well behind G-Retriever and RAG. RAG is the strongest overall (93.38 vs.\ 89.71), a gap attributable to information loss at triple extraction. Per-category results reveal substantial heterogeneity; notably, G-Retriever overtakes RAG on Gastronom\'ia (93.04 vs.\ 87.37), where relational abstraction filters out lexically crowded dense-retrieval noise.

\begin{table}[ht]
\centering
\small
\setlength{\tabcolsep}{4pt}
\resizebox{\columnwidth}{!}{%
\begin{tabular}{lrrcccc}
\toprule
Category & Zero-shot & RAG & Top-$k$ triples & G-Retriever \\
\midrule
Música       & 57.73 & \textbf{92.11} & 70.36 & 88.36 \\
Literatura   & 60.06 & \textbf{93.20} & 71.61 & 89.44\\
Cine         & 59.33 & \textbf{90.78} & 73.68 & 88.49\\
Folclore     & 63.82 & \textbf{96.69} & 78.78 & 92.87\\
Gastronomía  & 61.59 & 87.37 & 78.86 & \textbf{93.04} \\
Danza        & 63.22 & \textbf{96.03} & 73.01 & 93.12  \\
Pintura      & 59.56 & \textbf{90.97} & 74.36 & 88.81  \\
Artesanía    & 65.71 & \textbf{94.28} & 77.14 & 86.43 \\
\midrule
Global       & 60.17 & \textbf{93.38} & 75.60 & 89.71 \\
\bottomrule
\end{tabular}
}
\caption{Accuracy (\%) per category on LatamQA}
\label{tab:results-per-category} \vspace{-.3cm}
\end{table}

\subsection{Ablation Studies} 
\label{subsec:ablation}

Table~\ref{tab:results-qwen3b-merged} reports ablation results. 

\paragraph{Graph structure} 
Removing the trained projection (PCST as plain text, 73.10) falls below Top-$k$ triples, replicating the corresponding ablation of \citet{he2024}: the trained continuous prefix is the essential component. Interestingly, a linear projection matches or exceeds the graph encoder (88.90 vs.\ 86.13 without the textualized graph, 89.50 vs.\ 89.71 with it). We attribute this to the single-hop nature of the task: what the trained module supplies is a task-adapted continuous summary of the retrieved node and edge embeddings, functionally a form of prompt tuning \cite{Lester2021,Li2021g} conditioned on retrieved content, and message passing is an unnecessarily expressive way of producing it here.

\paragraph{Benchmark-aware extraction} 
Steering extraction with question-derived entity hints improves both Top-$k$ triples (+2.3 pp) and G-Retriever (+1.7 pp, reaching 91.41), confirming that standard extraction discards task-relevant content. The persistent gap with RAG (91.41 vs.\ 93.38) shows that extraction, not retrieval, is the main bottleneck.

\begin{table}[ht]
\centering
\small
\setlength{\tabcolsep}{6pt}
\resizebox{\columnwidth}{!}{%
\begin{tabular}{lcccc}
\toprule
Method & Encoder & Text Graph & Bench KGGen & Accuracy \\
\midrule
PCST & -- & \cmark & \xmark & 73.10 \\
Top-$k$ triples & -- & \cmark & \xmark & 75.60 \\
\midrule
PCST & GraphEnc & \xmark & \xmark & 86.13 \\
PCST & Linear   & \xmark & \xmark & 88.90 \\
PCST & Linear   & \cmark & \xmark & 89.50 \\
PCST & GraphEnc & \cmark & \xmark & 89.71 \\
\midrule
Top-$k$ triples & -- & \cmark & \cmark & 77.92 \\
PCST & GraphEnc & \cmark & \cmark & \textbf{91.41} \\
\bottomrule
\end{tabular}
}
\caption{Accuracy under different KGGen selection methods, graph encoders, and
benchmark settings. GraphEnc denotes the trained graph encoder of G-Retriever, Linear denotes mean-pooled node embeddings with a single projection. Bench KGGen refers to a KG constructed with LatamQA-specific
relations/entities.} \vspace{-.3cm}
\label{tab:results-qwen3b-merged}
\end{table}

\subsection{Multilingual transfer}
\label{sec:multilingual}
The five checkpoints from the Spanish cross-validation folds are applied without
further training to a Portuguese KG built from 397 Literatura articles, with
questions, triples and generation all in Portuguese. Since the projection aligns
a pooled subgraph representation with the reader's input space rather than
modeling any particular language, and since the encoder is multilingual
\cite{sturua2024a}, the mapping should be largely language-agnostic.
G-Retriever reaches 91.18, above its 89.44 on Spanish Literatura, while the
zero-shot baseline drops from 60.06 to 53.40 (Table~\ref{tab:results-qwen3b-mling-pt}).
A single trained projection can therefore serve several languages, provided the
retrieval backbone is itself multilingual.

\begin{table}[ht]
\centering
\small
\setlength{\tabcolsep}{6pt}
\begin{tabular}{llcc}
\toprule
Method & Accuracy  \\
\midrule
Zero-shot & $53.40$ \\
RAG                          & $93.20$ \\
 Top-$k$ triples              & $75.06$   \\
 PCST + GraphEnc (G-Retriever)  & $91.18$\\
\bottomrule
\end{tabular}
\caption{Zero-shot cross-lingual transfer on the Portuguese Literatura graph (397 articles).} \vspace{-.3cm}
\label{tab:results-qwen3b-mling-pt}
\end{table}

\subsection{Inference efficiency}
Graph-RAG is also lighter at inference: subgraph retrieval returns a compact set
of triples rather than full passages, reducing the average context from $2814$
to $875$ tokens (Table~\ref{tab:cost_comparison}). The cost moves offline: KG
extraction cost 72.26~EUR in API calls and training the graph encoder took
$\sim$9{,}942~s per fold (Appendix~\ref{app:cost}).

\begin{table}[ht]
\centering
\small
\setlength{\tabcolsep}{4pt}
\resizebox{\columnwidth}{!}{%
\begin{tabular}{lcc}
\toprule
\textbf{Cost dimension} & \textbf{RAG} & \textbf{Graph-RAG} \\
\midrule
Inference context (tokens) & $2814 \pm 223$ & $875 \pm 174$ \\
Storage footprint & 56.66\,MB & 36.20\,MB \\
\bottomrule
\end{tabular}
}
\caption{Inference-time cost comparison.} \vspace{-.3cm}
\label{tab:cost_comparison}
\end{table}

\section{Conclusion}
We benchmarked Graph-RAG against RAG on a culturally grounded, single-hop MCQ
dataset spanning eight Latin American thematic categories. G-Retriever reduces
base LLM error by 74\% on a 3.2$\times$ shorter context, and a benchmark-aware
KG narrows the residual gap with RAG to 2 pp, confirming extraction quality
rather than retrieval as the bottleneck; the projection also transfers
zero-shot to Portuguese. Both LatamQA and our KGs derive from Spanish
Wikipedia, whose coverage skews toward documented over orally transmitted
knowledge, and addressing this bias will require participatory sources
\cite{Zhou2025}.



\section{Limitations}



\paragraph{Evaluation format}
The four-way multiple-choice format admits a 25\% chance floor and lets a system succeed by elimination rather than recall: a partially relevant triple may suffice to discard distractors without stating the answer itself \cite{balepur-etal-2024-easy}. Extraction-induced information loss is therefore penalised only when it removes discriminative content, so the gap we report between flat retrieval and graph-based augmentation is likely a lower bound on the true cost of converting text into triples.

\paragraph{Single reader}
All results use one small reader, Qwen2.5-3B-Instruct, chosen to leave headroom for retrieval to matter. The G-Retriever projection is trained to align with this specific frozen reader, and we do not test whether the learned mapping transfers across backbones.

\paragraph{Narrow cross-lingual evidence}
Our multilingual transfer experiment covers a single category (Literatura) and a language typologically close to Spanish, on a graph an order of magnitude smaller than its Spanish counterpart, which likely eases retrieval. Broader claims would require matched-size graphs and typologically distant, lower-resource languages \cite{hu2020}.

\paragraph{Coverage bias}
Both LatamQA and our graphs are derived from Spanish Wikipedia, whose category distribution is highly uneven (1,306 articles for Música against 140 for Artesanía) and reflects editorial attention rather than cultural salience. Domains transmitted orally or through artisanal practice are underrepresented, so our benchmark measures Latin American culture as Wikipedia records it.

\bibliography{custom,JRC,Bias}

@article{Zhou2025,
    title = {{Disparities in LLM Reasoning Accuracy and Explanations: A Case Study on African American English}},
    year = {2025},
    author = {Zhou, Runtao and Wan, Guangya and Gabriel, Saadia and Li, Sheng and Gates, Alexander J and Sap, Maarten and Hartvigsen, Thomas},
    url = {http://arxiv.org/abs/2503.04099},
    arxivId = {2503.04099}
}

@inproceedings{Yang2026a,
    title = {{Fine-Tuning vs. RAG for Multi-Hop Question Answering with Novel Knowledge}},
    year = {2026},
    booktitle = {GEM 2026 @ ACL},
    author = {Yang, Zhuoyi and Song, Yurun and Ahmed, Iftekhar and Harris, Ian},
    pages = {384--392},
    url = {http://arxiv.org/abs/2601.07054},
    arxivId = {2601.07054}
}

@article{Ovadia2024,
    title = {{Fine-Tuning or Retrieval? Comparing Knowledge Injection in LLMs}},
    year = {2024},
    journal = {EMNLP 2024 - 2024 Conference on Empirical Methods in Natural Language Processing, Proceedings of the Conference},
    author = {Ovadia, Oded and Brief, Meni and Mishaeli, Moshik and Elisha, Oren},
    pages = {237--250},
    isbn = {9798891761643},
    doi = {10.18653/v1/2024.emnlp-main.15},
    arxivId = {2312.05934}
}

@article{Velickovic2018,
    title = {{Graph attention networks}},
    year = {2018},
    journal = {6th International Conference on Learning Representations, ICLR 2018 - Conference Track Proceedings},
    author = {Veli{\v{c}}kovi{\'{c}}, Petar and Casanova, Arantxa and Li{\`{o}}, Pietro and Cucurull, Guillem and Romero, Adriana and Bengio, Yoshua},
    pages = {1--12},
    doi = {10.1007/978-3-031-01587-8{\_}7},
    arxivId = {1710.10903}
}

@article{Yun2019,
    title = {{Graph transformer networks}},
    year = {2019},
    journal = {Advances in Neural Information Processing Systems},
    author = {Yun, Seongjun and Jeong, Minbyul and Kim, Raehyun and Kang, Jaewoo and Kim, Hyunwoo J.},
    number = {NeurIPS},
    volume = {32},
    issn = {10495258},
    arxivId = {1911.06455}
}

@inproceedings{He2024,
    title = {{MA-LMM: Memory-Augmented Large Multimodal Model for Long-Term Video Understanding}},
    year = {2024},
    booktitle = {CVPR},
    author = {He, Bo and Li, Hengduo and Jang, Young Kyun and Jia, Menglin and Cao, Xuefei and Shah, Ashish and Shrivastava, Abhinav and Lim, Ser-Nam},
    pages = {13504--13514},
    url = {http://arxiv.org/abs/2404.05726},
    arxivId = {2404.05726}
}

@inproceedings{Li2021g,
    title = {{Prefix-tuning: Optimizing continuous prompts for generation}},
    year = {2021},
    booktitle = {ACL-IJCNLP 2021 - 59th Annual Meeting of the Association for Computational Linguistics and the 11th International Joint Conference on Natural Language Processing, Proceedings of the Conference},
    author = {Li, Xiang Lisa and Liang, Percy},
    pages = {4582--4597},
    isbn = {9781954085527},
    doi = {10.18653/v1/2021.acl-long.353},
    arxivId = {2101.00190}
}

@article{Lester2021,
    title = {{The Power of Scale for Parameter-Efficient Prompt Tuning}},
    year = {2021},
    author = {Lester, Brian and Al-Rfou, Rami and Constant, Noah},
    url = {http://arxiv.org/abs/2104.08691},
    doi = {10.18653/v1/2021.emnlp-main.243},
    arxivId = {2104.08691}
}

@article{Mallen2023,
    title = {{When Not to Trust Language Models: Investigating Effectiveness of Parametric and Non-Parametric Memories}},
    year = {2023},
    journal = {Proceedings of the Annual Meeting of the Association for Computational Linguistics},
    author = {Mallen, Alex and Asai, Akari and Zhong, Victor and Das, Rajarshi and Khashabi, Daniel and Hajishirzi, Hannaneh},
    number = {Section 6},
    pages = {9802--9822},
    volume = {1},
    isbn = {9781959429722},
    doi = {10.18653/v1/2023.acl-long.546},
    issn = {0736587X},
    arxivId = {2212.10511}
}

@article{Hu2020,
    title = {{XTREME: A Massively Multilingual Multi-task Benchmark for Evaluating Cross-lingual Generalization}},
    year = {2020},
    author = {Hu, Junjie and Ruder, Sebastian and Siddhant, Aditya and Neubig, Graham and Firat, Orhan and Johnson, Melvin},
    url = {http://arxiv.org/abs/2003.11080},
    arxivId = {2003.11080}
}

@inproceedings{ranaldi-etal-2025-improving-multilingual,
    title = "Improving Multilingual Retrieval-Augmented Language Models through Dialectic Reasoning Argumentations",
    author = "Ranaldi, Leonardo  and
      Ranaldi, Federico  and
      Zanzotto, Fabio Massimo  and
      Haddow, Barry  and
      Birch, Alexandra",
    editor = "Christodoulopoulos, Christos  and
      Chakraborty, Tanmoy  and
      Rose, Carolyn  and
      Peng, Violet",
    booktitle = "Proceedings of the 2025 Conference on Empirical Methods in Natural Language Processing",
    month = nov,
    year = "2025",
    address = "Suzhou, China",
    publisher = "Association for Computational Linguistics",
    url = "https://aclanthology.org/2025.emnlp-main.461/",
    doi = "10.18653/v1/2025.emnlp-main.461",
    pages = "9064--9085",
    ISBN = "979-8-89176-332-6"
}

@inproceedings{balepur-etal-2024-easy,
    title = "It{'}s Not Easy Being Wrong: Large Language Models Struggle with Process of Elimination Reasoning",
    author = "Balepur, Nishant  and
      Palta, Shramay  and
      Rudinger, Rachel",
    editor = "Ku, Lun-Wei  and
      Martins, Andre  and
      Srikumar, Vivek",
    booktitle = "Findings of the Association for Computational Linguistics: ACL 2024",
    month = aug,
    year = "2024",
    address = "Bangkok, Thailand",
    publisher = "Association for Computational Linguistics",
    url = "https://aclanthology.org/2024.findings-acl.604/",
    doi = "10.18653/v1/2024.findings-acl.604",
    pages = "10143--10166"
}

@misc{karmim2026,
  title = {Leveraging {{Wikidata}} for {{Geographically Informed Sociocultural Bias Dataset Creation}}: {{Application}} to {{Latin America}}},
  shorttitle = {Leveraging {{Wikidata}} for {{Geographically Informed Sociocultural Bias Dataset Creation}}},
  author = {Karmim, Yannis and Pino, Renato and Contreras, Hernan and Lira, Hernan and Cifuentes, Sebastian and Escoffier, Simon and Mart{\'i}, Luis and Seddah, Djam{\'e} and Barri{\`e}re, Valentin},
  year = 2026,
  month = mar,
  number = {arXiv:2603.10001},
  eprint = {2603.10001},
  primaryclass = {cs},
  publisher = {arXiv},
  doi = {10.48550/arXiv.2603.10001},
  urldate = {2026-04-16},
  archiveprefix = {arXiv}
}

@inproceedings{yih2016,
  title = {The {{Value}} of {{Semantic Parse Labeling}} for {{Knowledge Base Question Answering}}},
  booktitle = {Proceedings of the 54th {{Annual Meeting}} of the {{Association}} for {{Computational Linguistics}} ({{Volume}} 2: {{Short Papers}})},
  author = {Yih, Wen-tau and Richardson, Matthew and Meek, Chris and Chang, Ming-Wei and Suh, Jina},
  editor = {Erk, Katrin and Smith, Noah A.},
  year = 2016,
  month = aug,
  pages = {201--206},
  publisher = {Association for Computational Linguistics},
  address = {Berlin, Germany},
  doi = {10.18653/v1/P16-2033},
  urldate = {2026-07-30}
}

@misc{baek2023,
      title={Knowledge-Augmented Language Model Prompting for Zero-Shot Knowledge Graph Question Answering}, 
      author={Jinheon Baek and Alham Fikri Aji and Amir Saffari},
      year={2023},
      eprint={2306.04136},
      archivePrefix={arXiv},
      primaryClass={cs.CL},
      url={https://arxiv.org/abs/2306.04136}, 
}

@misc{hudson2019,
      title={GQA: A New Dataset for Real-World Visual Reasoning and Compositional Question Answering}, 
      author={Drew A. Hudson and Christopher D. Manning},
      year={2019},
      eprint={1902.09506},
      archivePrefix={arXiv},
      primaryClass={cs.CL},
      url={https://arxiv.org/abs/1902.09506}, 
}

@misc{saha2021,
      title={ExplaGraphs: An Explanation Graph Generation Task for Structured Commonsense Reasoning}, 
      author={Swarnadeep Saha and Prateek Yadav and Lisa Bauer and Mohit Bansal},
      year={2021},
      eprint={2104.07644},
      archivePrefix={arXiv},
      primaryClass={cs.CL},
      url={https://arxiv.org/abs/2104.07644}, 
}

@misc{mo2025,
  title = {{{KGGen}}: {{Extracting Knowledge Graphs}} from {{Plain Text}} with {{Language Models}}},
  shorttitle = {{{KGGen}}},
  author = {Mo, Belinda and Yu, Kyssen and Kazdan, Joshua and Cabezas, Joan and Mpala, Proud and Yu, Lisa and Cundy, Chris and Kanatsoulis, Charilaos and Koyejo, Sanmi},
  year = 2025,
  date = {2025-11-06},
  eprint = {2502.09956},
  eprinttype = {arXiv},
  eprintclass = {cs},
  doi = {10.48550/arXiv.2502.09956},
  pubstate = {prepublished}
}

@misc{qwen2025,
  title = {Qwen2.5 {{Technical Report}}},
  author = {Qwen and Yang, An and Yang, Baosong and Zhang, Beichen and Hui, Binyuan and Zheng, Bo and Yu, Bowen and Li, Chengyuan and Liu, Dayiheng and Huang, Fei and Wei, Haoran and Lin, Huan and Yang, Jian and Tu, Jianhong and Zhang, Jianwei and Yang, Jianxin and Yang, Jiaxi and Zhou, Jingren and Lin, Junyang and Dang, Kai and Lu, Keming and Bao, Keqin and Yang, Kexin and Yu, Le and Li, Mei and Xue, Mingfeng and Zhang, Pei and Zhu, Qin and Men, Rui and Lin, Runji and Li, Tianhao and Tang, Tianyi and Xia, Tingyu and Ren, Xingzhang and Ren, Xuancheng and Fan, Yang and Su, Yang and Zhang, Yichang and Wan, Yu and Liu, Yuqiong and Cui, Zeyu and Zhang, Zhenru and Qiu, Zihan},
  year = 2025,
  month = jan,
  number = {arXiv:2412.15115},
  eprint = {2412.15115},
  primaryclass = {cs.CL},
  publisher = {arXiv},
  doi = {10.48550/arXiv.2412.15115},
  urldate = {2026-07-30},
  archiveprefix = {arXiv}
}

@misc{wang2024b1,
  title = {Multilingual {{E5 Text Embeddings}}: {{A Technical Report}}},
  shorttitle = {Multilingual {{E5 Text Embeddings}}},
  author = {Wang, Liang and Yang, Nan and Huang, Xiaolong and Yang, Linjun and Majumder, Rangan and Wei, Furu},
  year = 2024,
  month = feb,
  number = {arXiv:2402.05672},
  eprint = {2402.05672},
  primaryclass = {cs.CL},
  publisher = {arXiv},
  doi = {10.48550/arXiv.2402.05672},
  urldate = {2026-07-30},
  archiveprefix = {arXiv}
}

@misc{sturua2024a,
  title = {Jina-Embeddings-v3: {{Multilingual Embeddings With Task LoRA}}},
  shorttitle = {Jina-Embeddings-V3},
  author = {Sturua, Saba and Mohr, Isabelle and Akram, Mohammad Kalim and G{\"u}nther, Michael and Wang, Bo and Krimmel, Markus and Wang, Feng and Mastrapas, Georgios and Koukounas, Andreas and Wang, Nan and Xiao, Han},
  year = 2024,
  month = sep,
  number = {arXiv:2409.10173},
  eprint = {2409.10173},
  primaryclass = {cs.CL},
  publisher = {arXiv},
  doi = {10.48550/arXiv.2409.10173},
  urldate = {2026-07-30},
  archiveprefix = {arXiv}
}

\appendix

\section{Implementation Details}
\label{appendix:implementation}

\subsection{Embedding and Query} \label{app:emb_query}

\paragraph{Embedding model}
All retrieval components rely on \texttt{jinaai/jina-embeddings-v3} \cite{sturua2024a}, a state-of-the-art multilingual encoder matching the Spanish of both the corpus and the questions. Its 8,192-token context window — against 512 for comparable encoders such as \texttt{multilingual-e5-large-instruct} \cite{wang2024b1} — is a deliberate design choice: it allows coarse retrieval units, up to entire articles, and thus lets us vary retrieval granularity while holding the encoder fixed. Queries and Passages were encoded using the task-specific LoRA adapters of the model. 

\paragraph{Query construction} 
For all retrieval methods, the query is the concatenation of the question and its four answer options, without any indication of which option is correct. In a multiple-choice setting, the question alone is often an underspecified retrieval cue: it may lack the named entities and surface forms that anchor the relevant subgraph, whereas these frequently appear in the options themselves. Including all four options symmetrically enriches the lexical and semantic signal available to the retriever while preserving the integrity of the task, since no option is privileged at retrieval time. 
This design also aligns the evaluation with the inference-time setting of the reader, which observes the question and all options jointly.

\subsection{Retrieval Method Hyperparameters Optimization}
\label{appendix:hyperopti}

We ran an exhaustive grid search for the three retrieval modes—TOP-K TRIPLES, RAG and GRAPH-RAG (PCST)—on a held-out subset of 500 questions
sampled uniformly at random from the benchmark.

For RAG we varied the chunk size (32--512 tokens) and the number of retrieved chunks $k \in \{5, 10, 25\}$; for top-$k$ triples, the number of retrieved triples $k \in \{3, \dots, 100\}$; for Graph-RAG, the node and edge budgets and the PCST edge cost $c_e \in \{0.01, 0.1, 0.5\}$.
For the Graph-RAG grid we searched over the PCST retriever alone, without the trained GNN-based soft-prompting module, in the same configuration as the ablation of Section~\ref{subsec:ablation}. 
This choice is dictated by cost: a full grid would have required retraining the GNN for every cell, at roughly $10^4$ seconds per fold (Table~\ref{tab:cost_comparison}). It rests on the assumption that the retrieval component and the trained projection module are approximately separable, i.e., that the ranking of subgraph budgets induced by retrieval quality is preserved when the projection module is added downstream. 

\subsection{G-Retriever}

\begin{itemize}
    \item \textbf{Subgraph Retrieval via PCST.} Following \cite{he2024}, we first encode node and edge textual attributes with a pretrained language model and compute their cosine similarity to the query embedding, which serves as node prizes and edge costs. We then solve the Prize-Collecting Steiner Tree (PCST) problem to retrieve a connected subgraph $S^* = (V^*, E^*)$ maximizing query relevance while penalizing edge costs:
    \begin{equation}
        S^* = \arg\max_{S \subseteq G} \sum_{v \in V} \text{prize}(v) - \sum_{e \in E} \text{cost}_e.
    \end{equation}
    We set $top\_k_{nodes} = 15$, $top\_k_{edges} = 20$ and $\text{cost}_e = 0.5$, hyperparameters optimized via grid search.

    \item \textbf{Graph Encoder.} To model the structure of the retrieved subgraph $S^*$, we depart from the Graph Attention Network used in \cite{he2024} and instead employ a Graph Transformer \cite{Yun2019}, implemented via multi-head Transformer convolution layers that incorporate edge features into the attention computation. Our encoder stacks $L = 4$ layers with 8 attention heads, residual connections, layer normalization and dropout ($p = 0.1$), operating on node and edge embeddings of dimension 1024. Node representations are then aggregated into a single graph token via mean pooling: $h_g = \text{POOL}(\text{GraphTransformer}_{\phi_1}(S^*)) \in \mathbb{R}^{d_g}$, with $d_g = 1024$.
    
    Each layer updates node representations through multi-head attention over neighboring nodes, where attention coefficients are conditioned on both node and edge embeddings:
    \begin{equation*}
        x_i' = W_1 x_i + \sum_{j \in \mathcal{N}(i)} \alpha_{ij} W_2 x_j
    \end{equation*}
    \begin{equation*}
       \alpha_{ij} = \text{softmax}_j \left( \frac{(W_3 x_i)^\top (W_4 x_j + W_5 e_{ij})}{\sqrt{d}} \right).
    \end{equation*}

    \item \textbf{Projection Layer and Prompt Tuning.} A multilayer perceptron aligns the graph token with the hidden space of the frozen LLM ($d_l = 1024$): $\hat{h}_g = \text{MLP}_{\phi_2}(h_g) \in \mathbb{R}^{d_l}$. This graph token acts as a soft prompt, prepended to the embeddings of the textualized subgraph and the query; while the LLM parameters $\theta$ remain frozen, gradients flow through $\hat{h}_g$, enabling the optimization of $\phi_1$ and $\phi_2$ by standard backpropagation \cite{he2024}.
\end{itemize}

\subsection{Cross-Validation}

All the experiments were run on the full dataset using a $k$-fold train-val-test cross-validation.

\section{KG Generation} \label{app:kggen}

\subsection{Base pipeline}

KGGen \cite{mo2025} extracts entities and relations from raw text via structured LLM prompting. Each Wikipedia article is segmented into chunks of 5{,}000 characters and processed independently, yielding a per-article graph. Article-level graphs are then aggregated by category, and entity and edge resolution is applied at category scale so that each thematic category is represented by a single unified KG. Extraction is carried out in Spanish using Mistral Small 3.2 (\texttt{Mistral-Small-3.2-24B-Instruct-2506}) accessed via API. All LLM calls are routed through LiteLLM. Each triple is traced throughout the resolution process, so that for any given article we can recover the full set of triples originating from it in the final graph; this provenance information provides the ground truth against which retrieval is evaluated.

\subsection{Benchmark-aware knowledge graph}

For each category, we additionally construct a \textit{benchmark-aware} augmented KG. Given the questions and answer options of LatamQA --- including distractors --- we extract the entities and relations required to discriminate between the candidate answers using Mistral Small 3.2. The resulting entity and relation hints are injected into the extraction prompt, steering the KG towards elements relevant to the downstream task rather than arbitrary factual content. Extraction otherwise follows the base pipeline, and entity and edge resolution is applied unchanged. The procedure yields a second graph per category aligned with the question distribution while relying on the same source articles. This graph is used as an upper-bound diagnostic: it is not a deployable configuration, as it presupposes access to the evaluation questions at construction time.


\section{QA Prompts}

\subsection{Vanilla Prompt}

The Vanilla prompt to answer the LatamQA MCQ is shown in Figure \ref{fig:prompt_vanilla}.

\begin{figure}[ht]
\centering
\begin{tcolorbox}[colback=orange!5!white, colframe=black, rounded corners, width=0.95\linewidth]


\small

\vspace{0.2em}

\texttt{Answer the following multiple-choice question based on your general knowledge.} \\

\texttt{Question: \textcolor{green!50!black}{\{question\}}} \\\\

\texttt{Options:} \\
\texttt{A) \textcolor{green!50!black}{\{options['A']\}}} \\
\texttt{B) \textcolor{green!50!black}{\{options['B']\}}} \\
\texttt{C) \textcolor{green!50!black}{\{options['C']\}}} \\
\texttt{D) \textcolor{green!50!black}{\{options['D']\}}} \\\\

\texttt{Answer (single letter):}

\end{tcolorbox}

\caption{Prompt used for the LatamQA benchmark.}
\label{fig:prompt_vanilla}
\end{figure}

\subsection{Triplet-enhanced Prompt}

The prompt using context extracted from the graph is shown in Figure \ref{fig:prompt}.

\begin{figure}[H]
\centering
\begin{tcolorbox}[colback=orange!5!white, colframe=black, rounded corners, width=0.95\linewidth]


\small

\vspace{0.2em}

\texttt{Here is some context that may help answer the following multiple-choice question.} \\

\texttt{If the context is irrelevant or unclear, rely on your general knowledge.} \\\\

\texttt{Context:} \\
\texttt{\textcolor{green!50!black}{\{context\_block\}}} \\\\

\texttt{Question: \textcolor{green!50!black}{\{question\}}} \\\\

\texttt{Options:} \\
\texttt{A) \textcolor{green!50!black}{\{options['A']\}}} \\
\texttt{B) \textcolor{green!50!black}{\{options['B']\}}} \\
\texttt{C) \textcolor{green!50!black}{\{options['C']\}}} \\
\texttt{D) \textcolor{green!50!black}{\{options['D']\}}} \\\\

\texttt{Answer (single letter):}

\end{tcolorbox}

\caption{Prompt used for the LatamQA benchmark. \texttt{context\_block} contains the extracted triplets using the Top-$k$ triplet or PCST.}
\label{fig:prompt}
\end{figure}

\section{Retrieval Oracle Analysis} \label{app:retrieval}

To separate retrieval error from reading error, we restrict the candidate pool to the gold article for each question (oracle condition). Table~\ref{tab:results-qwen3b-retrieval} shows that RAG benefits most (+3.3\%, 93.38 to 96.72), and Top-$k$ triples gains 2.2\%, whereas G-Retriever gains only 0.6\% --- within its cross-fold dispersion. G-Retriever's residual error thus stems from information lost during triple extraction, not from retrieval failure. The gap with RAG widens under the oracle condition (from 3.7 to 6.5\%), confirming that the ceiling of the graph representation is set by extraction, not retrieval.

\begin{table}[ht]
\centering
\small
\setlength{\tabcolsep}{6pt}
\resizebox{\columnwidth}{!}{%
\begin{tabular}{lcc}
\toprule
Method & Full retrieval & Per-article (oracle) \\
\midrule
RAG                          & 93.38 & 96.72 \\
Top-$k$ triples              & 75.60 & 77.84 \\
PCST + GraphEnc (emb.\ + prompt)  & $89.71 \pm 1.38$ & $90.27 \pm 0.46$ \\
\bottomrule
\end{tabular}
}
\caption{Full-graph vs.\ per-article (oracle) retrieval, accuracy (\%).}
\label{tab:results-qwen3b-retrieval}
\end{table}

\section{Inference Cost Details}
\label{app:cost}

The two pipelines distribute their cost very differently across the system lifecycle. The text-based RAG baseline concentrates its expense at inference time: retrieved passages are injected verbatim into the prompt, yielding an average context of $2814.3$ tokens, roughly $3.2\times$ longer than the $874.8$ tokens required by Graph-RAG. This reduction is a direct consequence of retrieval granularity: subgraph retrieval returns a compact set of triples rather than full passages, discarding surrounding prose that contributes tokens without contributing evidence. These figures depend on retrieval hyperparameters and are not intrinsic to either paradigm.

Conversely, Graph-RAG front-loads its cost into an offline construction phase that RAG does not incur. Extracting the knowledge graphs with KGGen \cite{mo2025} on a 51M-character dataset required 72.26~EUR in API calls, and training the Graph encoder took 9{,}942~s on average per fold.

\begin{table}[ht]
\centering
\small
\setlength{\tabcolsep}{4pt}
\resizebox{\columnwidth}{!}{%
\begin{tabular}{lcc}
\toprule
\textbf{Cost dimension} & \textbf{RAG} & \textbf{Graph-RAG} \\
\midrule
Inference context (tokens) & $2814.3 \pm 222.9$ & $874.8 \pm 174.3$ \\
Storage footprint & 56.66\,MB (CSV) & 36.20\,MB (PKL) \\
GraphEnc training time (s) & n.a. & 9{,}942 \\
Graph extraction cost (EUR) & n.a. & 72.26 \\
\bottomrule
\end{tabular}
}
\caption{Full cost comparison between RAG and Graph-RAG. Inference context is mean $\pm$ std over the evaluation set; GraphEnc training is averaged over $k$-fold runs.}
\label{tab:cost_comparison_full}
\end{table}

\end{document}